\documentclass{article}
\usepackage[inline]{enumitem}
\usepackage{graphicx}
\usepackage{amsmath}
\usepackage{tabularx}
\usepackage{makecell}
\usepackage[dvipsnames]{xcolor}
\usepackage[font=footnotesize,labelfont=bf]{caption}
\usepackage[draft]{hyperref}
\usepackage{natbib}
\usepackage{placeins}
\usepackage[T1]{fontenc}
\usepackage{CJKutf8}
\usepackage{wrapfig,lipsum,booktabs}
\usepackage[english]{babel}

\usepackage[accepted]{./sty/icml2017_arxiv}
\usepackage{multirow}
\usepackage{dblfloatfix}
\usepackage{slashbox}
\usepackage{makecell}
\usepackage{xcolor,colortbl}
\usepackage{amssymb}
\usepackage[cjk]{kotex}
\usepackage[normalem]{ulem}
\usepackage{multicol}
\usepackage{caption}
\usepackage{subcaption}

\newcommand{\rep}[2]{\textcolor{red}{\sout{#1}}\textcolor{blue}{#2}}

\icmltitlerunning{Sentence transition matrix: An efficient approach that preserves sentence semantics}

\begin{document}
	\begin{CJK}{UTF8}{mj}
		
		\twocolumn[
		\icmltitle{Sentence transition matrix: An efficient approach that preserves sentence semantics}
		
		\begin{icmlauthorlist}
			\icmlauthor{Myeongjun Jang}{Ko}
			\icmlauthor{Pilsung Kang}{Ko}
		\end{icmlauthorlist}
		
		\icmlaffiliation{Ko}{School of Industrial Management Engineering, Korea University, Seoul, South Korea}
		\icmlcorrespondingauthor{Myeongjun Jang}{xkxpa@korea.ac.kr}
		\icmlcorrespondingauthor{Pilsung Kang}{pilsung\_kang@korea.ac.kr}
		
		\vskip 0.3in
		]
		\printAffiliationsAndNotice{}  

\begin{abstract}
Sentence embedding is a significant research topic in the field of natural language processing (NLP). Generating sentence embedding vectors reflecting the intrinsic meaning of a sentence is a key factor to achieve an enhanced performance in various NLP tasks such as sentence classification and document summarization. Therefore, various sentence embedding models based on supervised and unsupervised learning have been proposed after the advent of researches regarding the distributed representation of words. They were evaluated through semantic textual similarity (STS) tasks, which measure the degree of semantic preservation of a sentence and neural network-based supervised embedding models generally yielded state-of-the-art performance. However, these models have a limitation in that they have multiple parameters to update, thereby requiring a tremendous amount of labeled training data. In this study, we propose an efficient approach that learns a transition matrix that refines a sentence embedding vector to reflect the latent semantic meaning of a sentence. The proposed method has two practical advantages; (1) it can be applied to any sentence embedding method, and (2) it can achieve robust performance in STS tasks irrespective of the number of training examples.

\end{abstract}

\section{Introduction}
Sentence embedding, the task of transforming a sequence of words in a sentence into a fixed-dimensional vector form reflecting the intrinsic meaning, plays an important role in the field of natural language processing (NLP). It can be considered as an essential preprocessing step that transforms unstructured textual data into structured and continuous-valued vectors that can be used as input to machine learning algorithms to conduct various NLP tasks such as machine translation \cite{sutskever2014sequence,wu2016google}, document classification \cite{kim2014convolutional,conneau2017very}, and sentence matching \cite{hu2014convolutional,wan2016deep}. Because performances of many NLP tasks heavily rely on the word/sentence/document embedding methods, a large number of studies have been conducted since the advent of Doc2vec \cite{Doc2vec}, and progressive performance improvements have been reported whenever a new embedding method was proposed.

A good sentence embedding model should yield a vector value that closely captures the intrinsic semantic meaning of the sentence. Therefore, diverse experiments predicting the similarity score of two embedded sentences have been designed to evaluate how well the embedding model satisfies such requirements. These experiments were conducted in many researches, such as SIF \cite{SIF}, Sent2vec \cite{Sent2vec}, InferSent \cite{InferSent}, and FastSent \cite{FastSent}, and models trained through supervised learning generally showed far better performances than those based on unsupervised learning. However, supervised models have a limitation that they require a sufficient number of labeled training data to learn the model. For instance, Stanford Natural Language Inference (SNLI) dataset, which contains a collection of 570k manually labeled human-written English sentence pairs \cite{SNLI}, was used to train the InferSent model. Because it is an English dataset, one must construct a new dataset to train the embedding model for other languages, e.g., Russian, Korean, and Japanese, which is a heavy time- and resource-consuming work. Therefore, to increase the general usability of an embedding model, it is necessary to reduce reliance on labeled training data while closely preserving the intrinsic meaning of the sentence.

In this study, motivated by the studies that obtained successful results in cross-lingual embedding \cite{mikolov2013exploiting} and word-level translation \cite{smith2017offline} by training a simple transform matrix, we propose an approach to learn the transition matrix that refines the sentence vector generated by other sentence embedding models. The contributions of this research are as follows:
\begin{itemize}
	\item We propose an efficient method that successfully reflects sentence semantics through training the transition matrix. 
	\item The proposed method is independent of the size of training data and can be applied to any kind of sentence embedding model.
\end{itemize}
To reflect the intrinsic meaning of a sentence completely, we define a fundamental property that a good sentence embedding model should satisfy: the property of semantic coherence \cite{jang2018paraphrase}, which implies that paraphrase sentences should be closely located to each other in the sentence embedding space. Next, we derive an objective function to learn the transition matrix to closely satisfy the formulated property and train the matrix using the MSCOCO caption dataset. Experimental results show that the proposed approach is practically advantageous in that it outperforms existing sentence embedding models in various semantic textual similarity (STS) tasks.  

The rest of this paper is organized as follows. In Section 2, we briefly review the past researches on sentence embedding. In Section 3, we introduce the objective function for training the transition matrix and the training method. In Section 4, the effectiveness of the proposed method is demonstrated with the experimental results on STS tasks, along with their performance comparison with benchmark models. Finally, in Section 5, we conclude our present study, which leads us to some future research directions.

\section{Related work}
Recent researches on sentence embedding models are diverse for models based on unsupervised learning to those based on supervised learning. Unsupervised embedding models can be divided into two categories based on whether sentence sequence information (not word sequence information within a sentence) is required during training.

Doc2vec \cite{Doc2vec} is a representative model that does not need sentence sequence information. Paragraph vectors-distributed bag of words (PV-DBOW) and paragraph vector-distributed memory (PV-DM), two distinct learning methods of Doc2vec, train sentence vectors based on the same objective: maximizing the probability to predict words constituting the sentence. The probability is defined as the dot product between a sentence vector and a word vector. PV-DM considers sequential information of words by employing a moving window. In this method, a sentence vector is learned to predict a word appearing after the moving window using the words within the window and the sentence vector. However, in the PV-DBOW method, words included in the window are arbitrarily selected. Therefore, this is incapable of reflecting sequential information of words in a sentence.

\citet{hill2016learning} proposed the sequential denoising autoencoder (SDAE), which slightly corrupts the input sentence by adding random noise. The noise is added in two different ways. First, each word is randomly dropped in a sequence according to the probability $p_0$. Next, for the bigrams that are not overlapped, the order of two adjacent words is permuted according to the probability $p_x$. Then, the embedding model, which is a recurrent neural network (RNN) with a long short-term memory \cite{LSTM} (LSTM) cell, updates its parameter to generate the original sentence from the corrupted one. If $p_0 = p_x = 0$, the model becomes a sequential autoencoder (SAE), which does not add noise to the input. \citet{hill2016learning} also proposed a variant of the SDAE model that employs fixed pre-trained word vectors. This model is notated as ``S(D)AE + embs'' in our study.

\citet{SIF} proposed a simple embedding model named SIF, which computes a sentence vector as a weighted average of fixed pre-trained word embedding vectors. Despite its simplicity, SIF accomplished improved performance in STS tasks and outperformed many complex models based on RNNs if word weights are properly adjusted. Sent2vec \cite{Sent2vec} has a similar characteristic with SIF, which computes a sentence vector as a weighted average of word embedding vectors. However, Sent2vec trains not only the embedding vector of words that are unigram, but also that of n-gram tokens. It is different from SIF in that it employs n-gram embedding vectors to generate sentence vectors.

Contrary to previously demonstrated sentence embedding models that exploit information from corpora to learn sentence vectors, C-PHRASE \cite{kruszewski2015jointly} requires external information. C-PHRASE uses information from the syntactic parse tree of each sentence. This additional knowledge is included in the training objective of C-BOW \cite{Word2vec}.

SkipThought \cite{SkipThought} is a sentence embedding model in which sentence sequences are mandatory during the training. It has a sequence-to-sequence structure and expands the training objective of Skip-gram \cite{Word2vec} for learning word embedding vectors to a sentence level. Similar to the Skip-gram model, which updates word embedding vectors by predicting the surrounding words when the center word is given, the training objective of SkipThought is to generate the preceding and following sentences when a sentence is given.

FastSent \cite{FastSent}, similar to SkipThought, is a sentence embedding model aimed at predicting the surrounding sentences of a given sentence. FastSent learns the source word embedding $\textbf{u}_w$ and target word embedding $\textbf{v}_w$. When three consecutive sentences $S_{i-1}$, $S_i$, $S_{i+1}$ are given, $\textbf{s}_i$, the representation vector of $S_i$, is calculated as the sum of the source word embedding vectors:
\begin{gather}
\textbf{s}_i =\sum_{w \in \textbf{S}_i} \textbf{u}_w.
\end{gather}
Then, the cost function is simply defined as follows:
\begin{gather}
\sum_{w \in S_{i-1} \cup S_{i+1}} softmax(\textbf{s}_i \cdot \textbf{v}_w).
\end{gather}
In addition, \citet{FastSent} also proposed a variant model (FastSent+AE) that predicts not only the adjacent sentences but also the center sentence $S_i$. FastSent with such a simple structure\rep{,}{} takes much less training time than SkipThought.

Siamese C-BOW \cite{kenter2016siamese} shares a common concept with SIF and Sent2vec: defining a sentence vector as the average of word embedding vectors. In addition, it is similar to SkipThought and FastSent in that it is also trained to predict surrounding sentences when the center sentence is given. However, Siamese C-BOW employs a Siamese neural network \cite{koch2015siamese} structure, which is significantly different from the sentence embedding models described above.

InferSent \cite{InferSent} is a sentence embedding model trained through supervised tasks. Inspired by previous researches in computer vision, where a large number of models are pre-trained through a classification task based on the ImageNet \cite{Imagenet} dataset, \citet{InferSent} performed a research to determine the effectiveness of supervised tasks in the learning of a sentence embedding model in the field of NLP. Through experiments, \citet{InferSent} concluded that a sentence embedding model having a bidirectional LSTM structure, trained on the SNLI dataset, yielded state-of-the-art performance in various NLP tasks.

All sentence embedding models using sentence sequence information, described above, require a specific dataset to train the model. For instance, SkipThought and FastSent used the Book Corpus dataset \cite{Bookcorpus} and InferSent used the SNLI dataset to train the embedding models. In the case of sentence embedding models independent of sentence sequence information, a large document dataset, such as Wikipedia sentences, was also used. Furthermore, these models have a limitation in completely reflecting the traits of paraphrase sentences conveying the same meaning but having a different word usage, because these models generate sentence vectors based on word embedding vectors. In this study, we propose an efficient approach that successfully preserves sentence semantics by employing only a small amount of paraphrase sentences.

\section{An efficient transition matrix for generating sentence embedding vectors}
In this section, we first define a fundamental property that a good sentence embedding model should satisfy. Next, we describe the method to learn the transition matrix from the derived property. Finally, we demonstrate the processing method for training data when learning the transition matrix.

\subsection{Property of semantic coherence}
Semantic coherence refers to a fundamental property that a good sentence embedding model should satisfy: if two sentences have similar meaning, they should be located close to each other in the embedding space. The semantic coherence of a sentence embedding model can be evaluated as follows:

\vspace{5pt}
\noindent \textbf{Definition:} The degree of semantic coherence of a sentence embedding model is proportional to the similarity between the embedding vectors of paraphrase sentences generated by the sentence embedding model.
\vspace{5pt}

The above definition can be expressed mathematically. Assume that the set $\textbf{I}$ is a vector set of input sentences and the set $\textbf{P}$ is a vector set of paraphrase sentences corresponding to the input sentences:
\begin{equation}
\begin{gathered}
\textbf{I} = (\textbf{i}_1,\textbf{i}_2,...,\textbf{i}_n), ~~ 
\textbf{P} = (\textbf{p}_1,\textbf{p}_2,...,\textbf{p}_n),
\end{gathered}
\end{equation}
where $(\textbf{i}_k,\textbf{p}_k)$ is a pair of paraphrase sentence vectors. Then, the similarity matrix of elements belonging to each set can be computed as follows:
\begin{equation}
\begin{gathered}
\hat{\textbf{S}} = \textbf{I}\textbf{P}^T.
\end{gathered}
\end{equation}
The normalized similarity matrix $\textbf{N}$ can be obtained by the element-wise multiplication of $\textbf{N}$ with $\hat{\textbf{S}}$ as follows:
\begin{equation}
\begin{gathered}
\textbf{N}_{ij} = \frac{1}{||\textbf{i}_i||||\textbf{p}_j||}, ~~ 
\textbf{S} = \textbf{N}\times\hat{\textbf{S}}.
\end{gathered}
\end{equation}
The tupple $(\textbf{i}_i,\textbf{p}_j)$ is a pair of sentence vectors sharing a similar intrinsic meaning when $i=j$. However, when $i\neq j$, the sentence vectors have different meanings. Therefore, the similarity of two sentence vectors $\textbf{i}_i$ and $\textbf{p}_j$ should be close to 1 when $i=j$ and should be minimized when $i \neq j$. Hence, the similarity matrix $\textbf{S}$ of two vector sets $\textbf{I}$ and $\textbf{P}$ should satisfy the following two conditions if the sentence embedding model fulfills the property of semantic coherence.
\begin{center}
	\begin{tabular}{l}
		\textbf{Condition 1}: $\textbf{d}=diag(\textbf{S}) \approx \stackrel{\rightarrow}{1}$,\\
		\textbf{Condition 2}: $|\textbf{S}-diag(\textbf{d})|$ should be minimized.
	\end{tabular}
\end{center}

\subsection{Sentence transition matrix}
In this study, we attempted to devise an approach satisfying the semantic coherence by learning a minimal number of weights. To achieve this, we trained the transition matrix, which elaborates the pre-generated sentence vectors to meet the conditions described above. Assume that $\textbf{I}^\mathrm{M}$ and $\textbf{P}^\mathrm{M}$ refer to the sentence vector sets generated by the sentence embedding model $\mathrm{M}$ of the input sentence and the corresponding paraphrase sentences, respectively.
\begin{equation}
\begin{gathered}
\textbf{I}^\mathrm{M} = (\textbf{i}_1^\mathrm{M},\textbf{i}_2^\mathrm{M},...,\textbf{i}_n^\mathrm{M}), ~~ 
\textbf{P}^\mathrm{M} = (\textbf{p}_1^\mathrm{M},\textbf{p}_2^\mathrm{M},...,\textbf{p}_n^\mathrm{M}).
\end{gathered}
\end{equation}
The main purpose of this study is to train the transition matrix $\textbf{W}$ using $\textbf{I}^\mathrm{M}$ and $\textbf{P}^\mathrm{M}$. First, each set of sentence vectors is multiplied by the transition matrix as follows:
\begin{equation}
\begin{gathered}
\hat{\textbf{I}}^\mathrm{M} = \textbf{W}\textbf{I}^\mathrm{M}, ~~
\hat{\textbf{P}}^\mathrm{M} = \textbf{W}\textbf{P}^\mathrm{M}.
\end{gathered}
\end{equation}
Then, the similarity matrix of the sentence vectors to which the transition matrix is applied is computed as follows:
\begin{equation}
\begin{gathered}
\hat{\textbf{S}}^\mathrm{M} = \hat{\textbf{I}}^\mathrm{M} (\hat{\textbf{P}}^\mathrm{M})^T, ~~
\textbf{S}^\mathrm{M} = \textbf{N} \times \hat{\textbf{S}}^\mathrm{M}, \\ ~~ \text{where} ~ \textbf{N}_{ij} = \frac{1}{||\textbf{i}_i^\mathrm{M}|| || \textbf{p}_j^\mathrm{M}||}.
\end{gathered}
\end{equation}
Next, from the formulated conditions of the semantic coherence property, the diagonal and non-diagonal losses are defined as follows:
\begin{equation}
\begin{gathered}
\text{diagonal\_loss} = average(|diag(\textbf{S}^\mathrm{M} - \stackrel{\rightarrow}{1})|), \\
\text{non-diagonal\_loss} = average(|\textbf{S}^\mathrm{M} - diag(diag(\textbf{S}^\mathrm{M}))|).
\end{gathered}
\end{equation}
Finally, the final training loss is defined as follows:
\begin{equation}
\begin{aligned}
\text{loss} &= \lambda \times \text{non-diagonal\_loss} \\ &\quad + (1-\lambda) \times \text{diagonal\_loss},
\end{aligned}
\end{equation}
where $\lambda$ is the user-specific hyperparameter that controls the trade-off between the two losses. The sentence vector for sentence $x$ after training the transition matrix is computed as follows:
\begin{equation}
\begin{gathered}
\textbf{sv}_x = \textbf{W}\cdot\mathrm{M}(x),
\end{gathered}
\end{equation}
where $\mathrm{M}(x)$ is the vector value of the sentence $x$ generated by model $\mathrm{M}$.

Model $\mathrm{M}$ can be any kind of sentence embedding model. However, we defined the average of pre-trained word vectors as a sentence vector to show that the performance improvement could be obtained by a simple method. We used two kinds of pre-trained word vectors: GloVe vectors \cite{Glove} and Google Word2vec\footnote{\href{url}{https://code.google.com/archive/p/word2vec/}}.

\begin{figure*}[t!]
	\centering
	\includegraphics[width=\linewidth]{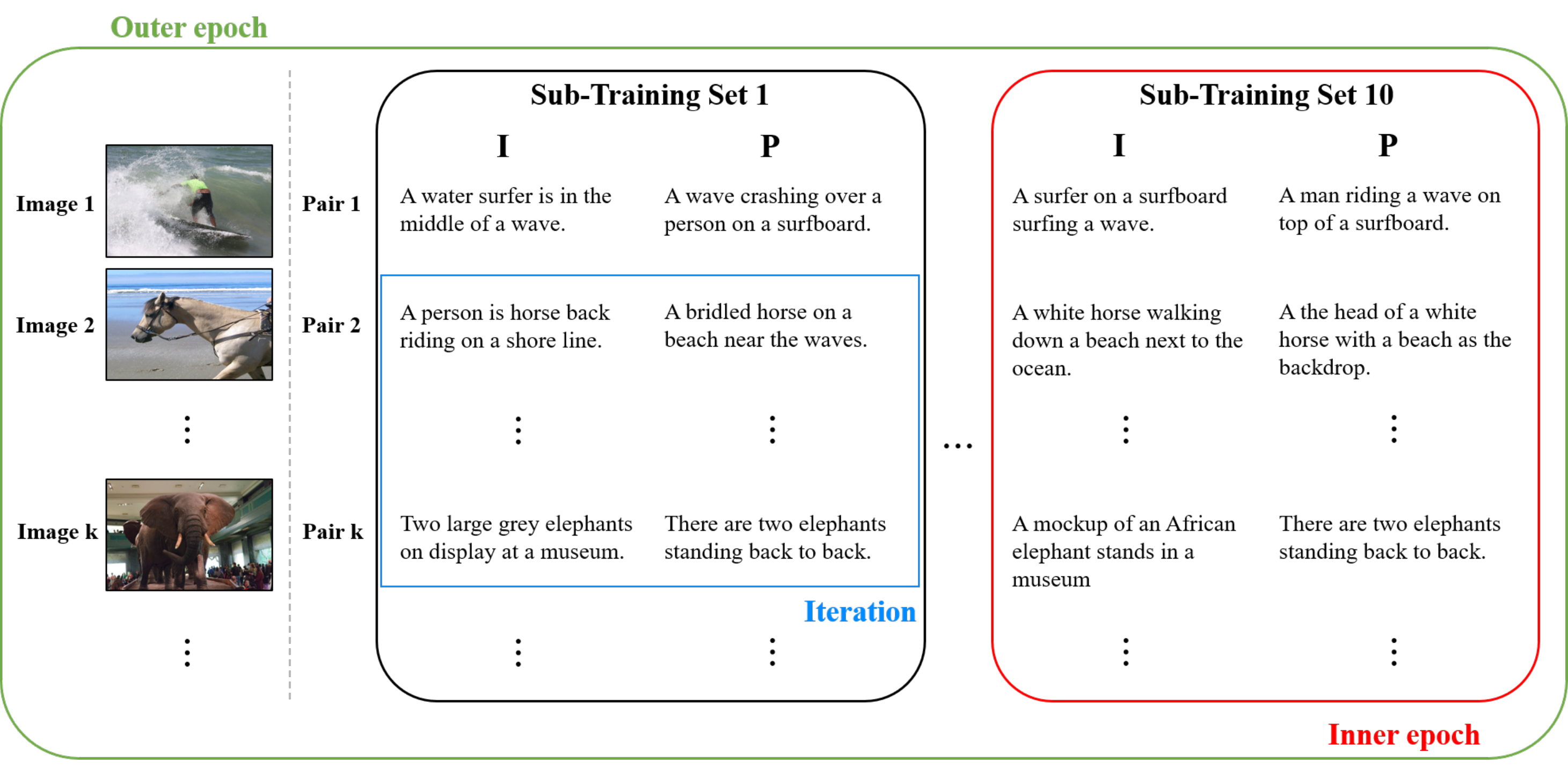}
	\caption{Description of training data composition}
	\label{fig:tr_data}
\end{figure*}

\subsection{Training transition matrix}
\noindent \textbf{Composing training dataset:} To train the transition matrix, we employed the MSCOCO 2017 training dataset \cite{MSCOCO}, which is widely used as a paraphrase dataset in many researches \cite{prakash2016neural,gupta2017deep}. This dataset contains a minimum of five captions per image for 118,284 images. The total number of unique captions is 591,753. To train the transition matrix proposed in this study, $\textbf{i}_i$ and $\textbf{p}_j$ should be mutually exclusive except for the case when $i=j$. In other words, in a mini-batch, only one paraphrase sentence should be included for one corresponding sentence when training the transition matrix. Hence, we constructed the training dataset as explained below.

First, note that it is possible to create ten unique sentence pairs for each image because at least five captions are provided for each image. Therefore, for all images, if we include only one sentence pair per image in one sub-training set, then the sub-training set does not contain sentence pairs with overlapping meaning and totally ten sub-training sets can be constructed. In our experiment, we denoted the training of one sub-training set as \textit{inner epoch} and of all sub-training sets as \textit{outer epoch}. The description of training data composition is provided in Figure \ref{fig:tr_data}.

\noindent \textbf{Training option:} Both GloVe vectors and Google Word2vec have word vectors of 300 dimensions. Therefore, the transition matrix has a size of $300 \times 300$. We employed Xavier initialization \cite{Xavier} and weight updates were performed with a mini-batch size of 512. We trained the transition matrix for five outer epochs using the RMSprop optimizer\footnote{\href{url}{http://www.cs.toronto.edu/~tijmen/csc321/slides/lecture\_slides\_lec6.pdf}}.

\section{Experiments}
\subsection{Textual similarity task}
\noindent \textbf{Data set:} To evaluate how well the proposed approach reflects the intrinsic meaning of sentences, we conducted STS tasks (2012--2016) \cite{STS12,STS13,STS14,STS15,STS2016} and SemEval 2014 semantic relatedness task (SICK) \cite{marelli2014semeval}. The objective of these tasks is to predict the similarity score of two given sentences. Their performance is evaluated by comparing the predicted score with the ground truth, which is the similarity score determined by human judgments. The evaluation metric is Pearson's $r$ \cite{pearson1895note} and Spearman's $\rho$ \cite{spearman1904proof}. Similar to previous researches, we defined the predicted similarity score as the cosine similarity between two sentence vectors.

\noindent \textbf{Experiment design:} We conducted the experiments in three different levels based on the number of training data to investigate the relationship between the performance and size of training data. The first level is the case of using all available data. The second and third levels are the cases where only 50 \% and 10 \%, respectively, of the total training data are used.

In the proposed approach, the hyperparameter $\lambda$ is a key value affecting the model performance. In this experiment, we empirically determined $\lambda$ from a sufficient number of candidate values. The $\lambda$ values for GloVe vectors and Google Word2vec are set to 0.7 and 0.9, respectively.

\subsection{Effect of the transition matrix on performance}
\begin{figure*}[t!]%
	\centering{
		\begin{tabular}{cc}
			\includegraphics[width=0.43\linewidth]{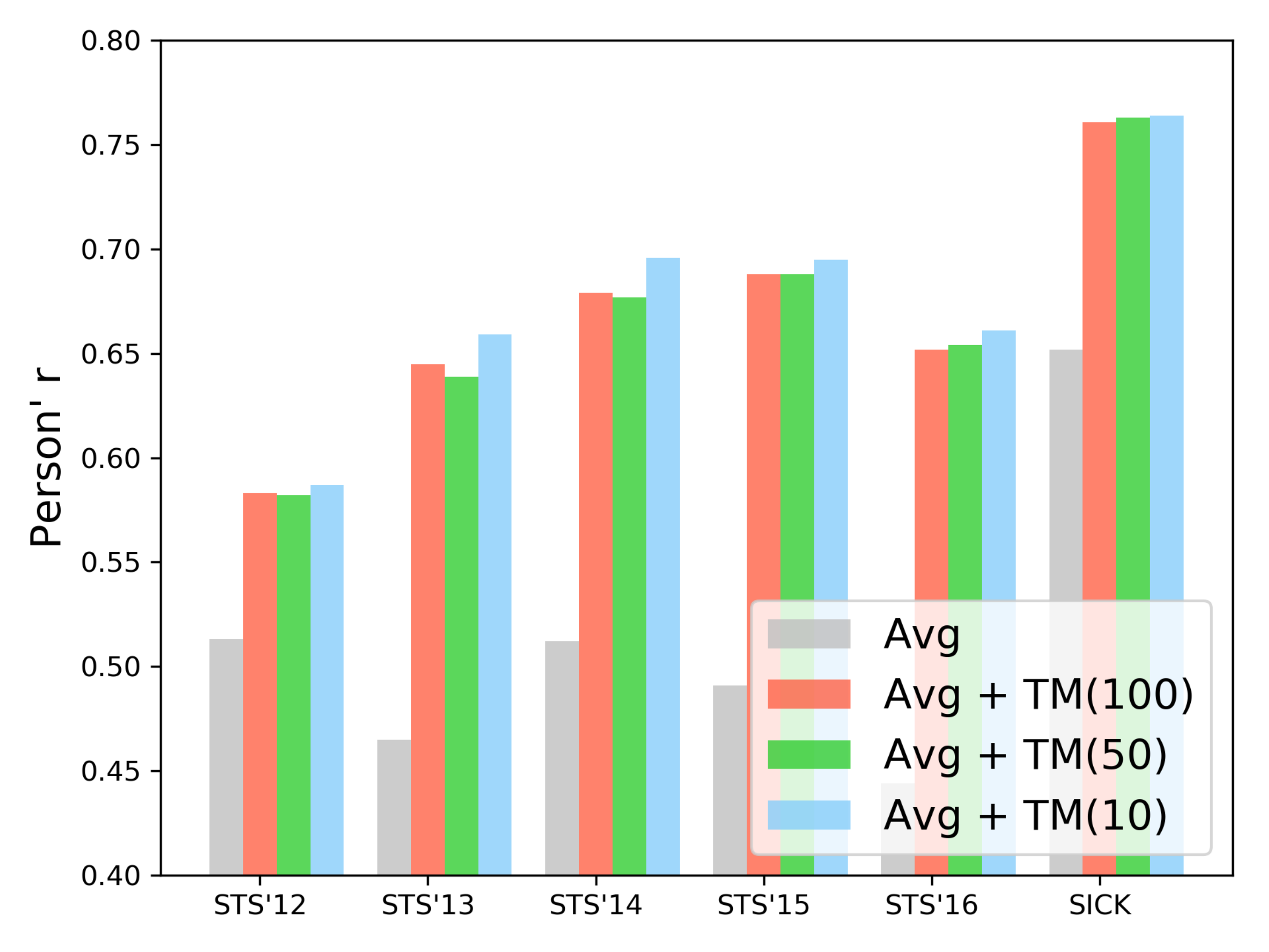} &
			\includegraphics[width=0.43\linewidth]{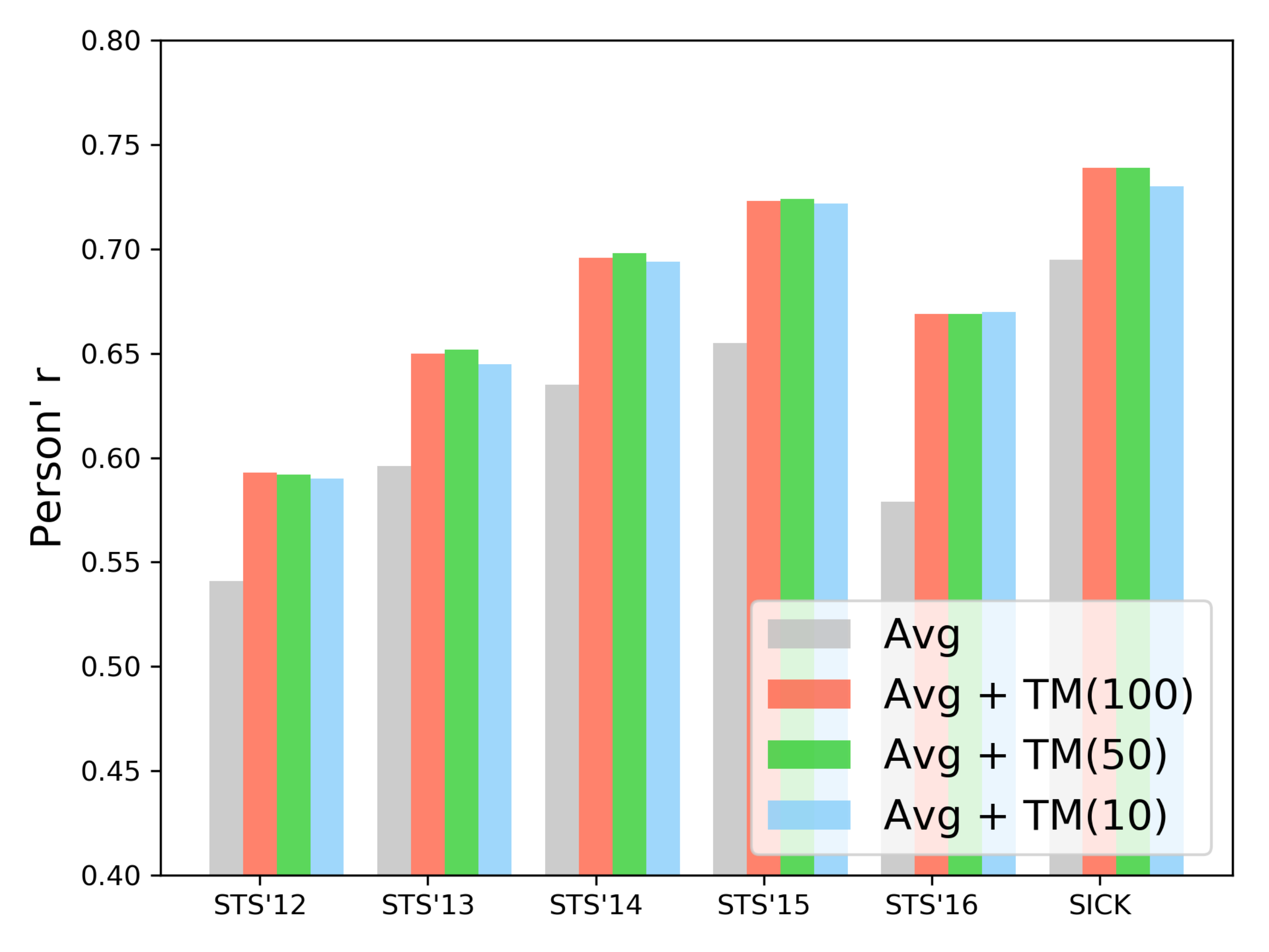} \\
			\footnotesize{(a) Glove} & \footnotesize{(b) Google Word2vec} \\			
			
	\end{tabular}} %
	\caption{Effect of transition matrix}
	\label{fig:visulaization}%
\end{figure*}

We first evaluated the effect of the transition matrix by comparing the performances before and after applying the transition matrix to sentence vectors that are defined as the average of pre-trained word vectors. The result is summarized in Table \ref{table1.Transition_Matrix_Performance} and Figure \ref{fig:visulaization}. The notation ``TM'' denotes that the transition matrix is applied, and the number in the parentheses indicates the percentage of training data used. The evaluation metric recorded in Table \ref{table1.Transition_Matrix_Performance} is Pearson's $r$.

\begin{table*}[t!]
	\begin{center}
		\caption{Result of semantic textual similarity tasks} \label{table1.Transition_Matrix_Performance}%
		\renewcommand{\arraystretch}{1.2}
		\centering{\setlength\tabcolsep{7pt}
			\begin{tabular}{cc*6c}
				\toprule
				\multicolumn{2}{c}{Model}  & {STS 12} & {STS 13} & {STS 14} & {STS 15} & {STS 16} & {SICK} \\ \hline
				\multirow{4}{*}{Glove} & Avg & 0.513 & 0.465 & 0.512 & 0.491 & 0.444 & 0.652 \\ 
				& Avg+TM(100) & 0.583 & 0.645 & 0.679 & 0.688 & 0.652 & 0.761 \\ 
				& Avg+TM(50)  & 0.582 & 0.639 & 0.677 & 0.688 & 0.654 & 0.763 \\ 
				& Avg+TM(10)  & \textbf{0.587} & \textbf{0.659} & \textbf{0.696} & \textbf{0.695} & \textbf{0.661} & \textbf{0.764} \\ 
				\hline
				\multirow{4}{*}{\makecell{Google \\Word2vec}} & Avg & 0.541 & 0.596 & 0.635 & 0.655 & 0.579 & 0.695 \\ 
				& Avg+TM(100) & \textbf{0.593} & 0.650 & 0.696 & 0.723 & 0.669 & \textbf{0.739} \\ 
				& Avg+TM(50)  & 0.592 & \textbf{0.652} & \textbf{0.698} & \textbf{0.724} & 0.669 & \textbf{0.739} \\ 
				& Avg+TM(10)  & 0.590 & 0.645 & 0.694 & 0.722 & \textbf{0.670} & 0.730 \\  \bottomrule
		\end{tabular}}
	\end{center}
\end{table*}

Experimental results show that the proposed transition matrix remarkably improves the performance in both the pre-trained word vectors. The training time was about 130, 60, and 13 s when 100, 50, and 10 \%, respectively, of the training data was used to train $\mathbf{W}$.In the case of Google Word2vec, the performance improvement is 5 $\sim$ 16 \%, while the transition matrix with Glove vector shows a very high performance improvement of 13 $\sim$ 49 \%. Both the transition matrices do not show significant performance differences according to the amount of training data. Therefore, it can be concluded that the performance of a transition matrix is independent of the amount of paraphrase data used in training.

\subsection{Comparison with other sentence embedding models}
After confirming the effectiveness of the transition matrix, we compared its performance with previously proposed sentence embedding models. The benchmark models are as follows:
\begin{itemize}
	\item \textbf{Embedding models using sentence sequence information}: S(D)AE, PV-DBOW, PV-DM, Sent2vec, C-PHRASE, and SIF.
	\item \textbf{Embedding models not using sentence sequence information}: SkipThought, FastSent, and Siamese C-BOW.
	\item \textbf{Supervised task-based sentence embedding model}: InferSent
\end{itemize}

\begin{table*}[t!]
	\begin{center}
		\caption{Result of semantic textual similarity tasks} \label{table2.Comparison}%
		\renewcommand{\arraystretch}{1.3}
		\footnotesize{
			\centering{\setlength\tabcolsep{3.5pt}
				\begin{tabular}{*9c}
					\toprule
					\multirow{2}{*}{Model} & \multicolumn{7}{c}{STS 2014} & {SICK}\\ 
					& News & Forum & WordNet & Twitter & Images & Headlines & Avg & Test+train\\ \hline	
					SAE &.17/.16 &.12/.12 &.30/.23 &.28/.22 &.49/.46 &.13/.11 &.27/.23 &.32/.31\\
					SAE+embs&.52/.54&.22/.23&.60/.55&.60/.60&.64/.64&.41/.41&.53/.52&.47/.49\\
					SDAE&.07/.04&.11/.13&.33/.24&.44/.42&.44/.38&.36/.36&.33/.30&.46/.46\\
					SDAE+embs&.51/.54&.29/.29&.56/.50&.57/.58&.59/.59&.43/.44&.51/.51&.46/.46\\
					PV-DBOW&.31/.34&.32/.32&.53/.50&.43/.46&.46/.44&.39/.41&.42/.43&.42/.46\\
					PV-DM&.42/.46&.33/.34&.51/.48&.54/.57&.32/.30&.46/.47&.44/.45&.44/.40\\
					C-BOW&.57/.61&.43/.44&.72/.69&.\textbf{\uline{71/.75}}&.71/.73&.55/.59&.64/.66&.60/.69\\
					Uni TF-IDF&.48/.48&.40/.38&.60/.59&.63/.65&.72/.74&.49/.49&.58/.58&.52/.58\\
					Sent2vec uni&.62/.67&\textbf{\uline{.49/.49}}&.75/.72&\textbf{.70/\uline{.75}}&\textbf{\uline{.78}}/.82&\textbf{.61/.63}&\textbf{\uline{.68/.70}}&.61/.70\\
					Sent2vec bi&.62/.67&\textbf{\uline{.49/.49}}&.71/.68&\textbf{.70/\uline{.75}}&.75/.79&.59/.62&\textbf{.66}/.69&\textbf{.62}/.70\\
					C-PHRASE&.69/.71&.43/.41&.76/.73&.60/.65&.75/.79&\textbf{.60/\uline{.65}}&\textbf{.66}/.68&.60/.72\\
					SIF (Glove)&-&-&-&-&-&-&- /.69&- /.72\\ \hline
					
					SkipThought&.44/.45&.14/.15&.39/.34&.42/.43&.55/.60&.43/.44&.42/.43&.57/.60\\
					FastSent&.58/.59&.41/.36&.74/.70&.63/.66&.74/.78&.57/.59&.64/.65&.61/.72\\
					FastSent+AE&.56/.59&.41/.40&.69/.64&.70/.74&.63/.65&.58/.60&.62/.65&.60/.65\\
					Siamse C-BOW&.58/.59&.42/.41&.66/.61&.71/.73&.63/.65&\textbf{\uline{.63}/.64}&.63/.63&-\\ \hline
					
					InferSent&-&-&-&-&-&-&\textbf{.67/\uline{.70}}&- \\ \hline
					
					Glove avg&.66/.66&.29/.22&.63/.56&.59/.57&.57/.58&.45/.46&.53/.51&.55/.65\\
					Google w2v avg&.64/.69&.31./31&.76/.73&.66/.70&.68/.71&.52/.58&.61/.64&.60/.70\\ \hline
					
					Glove avg+TM(100)&\textbf{.70/\uline{.73}}&.43/.44&.74/.73&.57/.64&\textbf{.77/\uline{.84}}&.58/\textbf{.63}&.64/.68&.61/\textbf{\uline{.76}}\\
					Glove avg+TM(50)&\textbf{.70/.72}&.43/.44&.73/.72&.58/.64&\textbf{.77}/.83&.58/\textbf{.63}&.64/.68&.61/\textbf{\uline{.76}}\\
					Glove avg+TM(10)&\textbf{\uline{.71/.73}}&\textbf{.45/.45}&.78/.77&.61/.67&\textbf{\uline{.78/.84}}&.58/\textbf{.63}&\textbf{.66/\uline{.70}}&\textbf{.62/\uline{.76}}\\
					Google w2v avg+TM(100)&.62/.69&.37/.38&\textbf{\uline{.81/.81}}&.66/.73&\textbf{.77}/.83&.54/.61&.65/\textbf{\uline{.70}}&\textbf{\uline{.63}}/.74\\
					Google w2v avg+TM(50)&.63/.69&.37/.38&\textbf{\uline{.81/.81}}&.66/.73&\textbf{.77/}\textbf{\uline{.84}}&.54/.61&.65/\textbf{\uline{.70}}&\textbf{\uline{.63}}/.74\\
					Google w2v avg+TM(10)&.63/.69&.36/.38&\textbf{\uline{.81}/.80}&.67/.73&\textbf{.77}/.83&.54/.60&.65/.69&\textbf{.62}/.73\\
					\bottomrule
		\end{tabular}}}
	\end{center}
\end{table*}

The tasks used for comparison are the STS 2014 and SICK relatedness tasks, both of which were used in the benchmark models. The benchmark results were obtained by \citet{Sent2vec}, \citet{SIF}, and \citet{InferSent}. For the models discussed above, the result of \citet{Sent2vec} contains the model defining a sentence vector as the simple average of word embeddings trained by the C-BOW \cite{Word2vec} method. It also includes the result of TF-IDF representation, which is the weighted word count of the 200,000 most common words. We underlined the best performance for the dataset, while the top three performances are shown in bold. The order of recording the results is Spearman's $\rho$ / Pearson's $r$. The results are summarized in Table \ref{table2.Comparison}. 

\begin{figure*}[t!]%
	\centering{
		\begin{tabular}{cc}
			\includegraphics[width=0.8\linewidth]{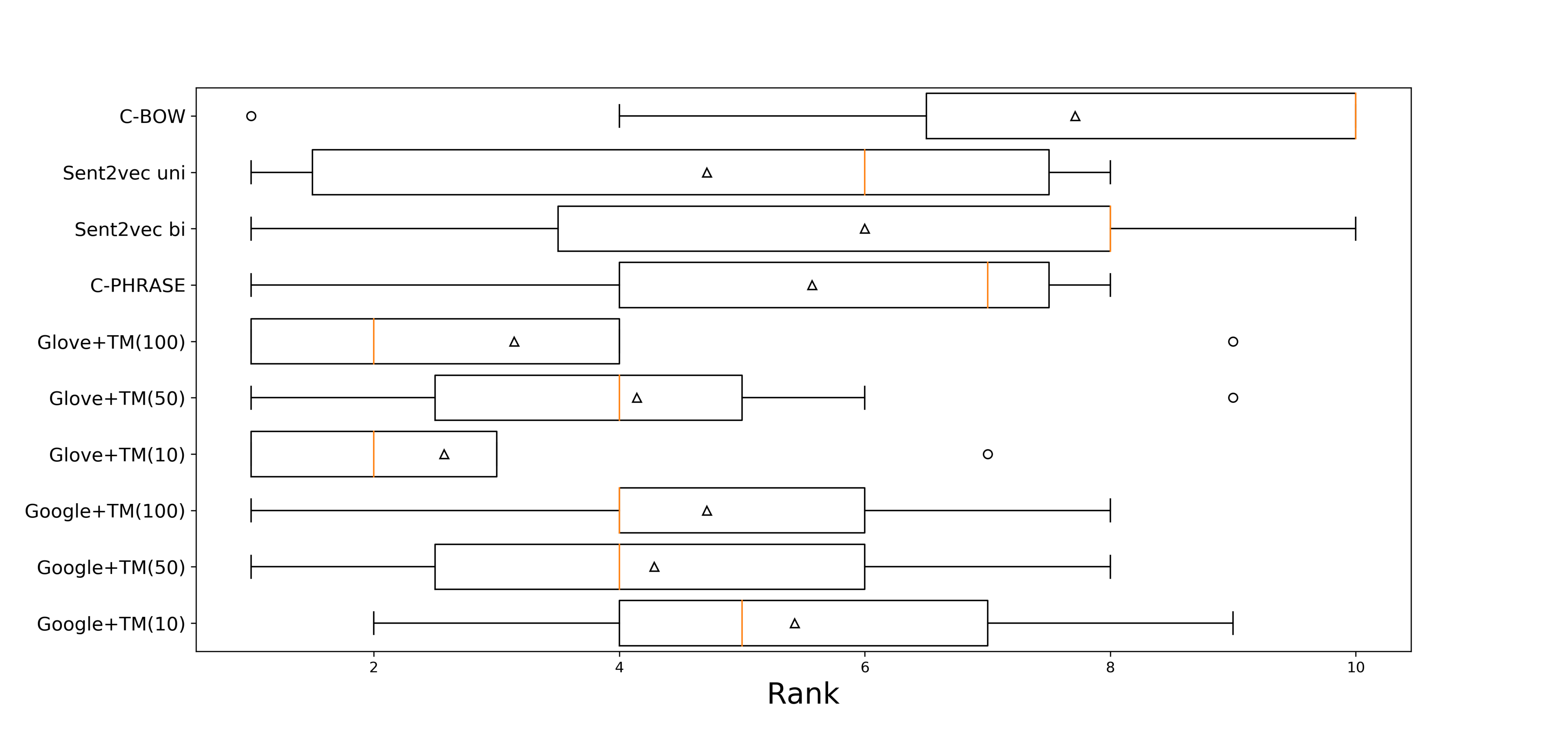} \\
			\footnotesize{(a) Pearson's $r$}\\
			\includegraphics[width=0.8\linewidth]{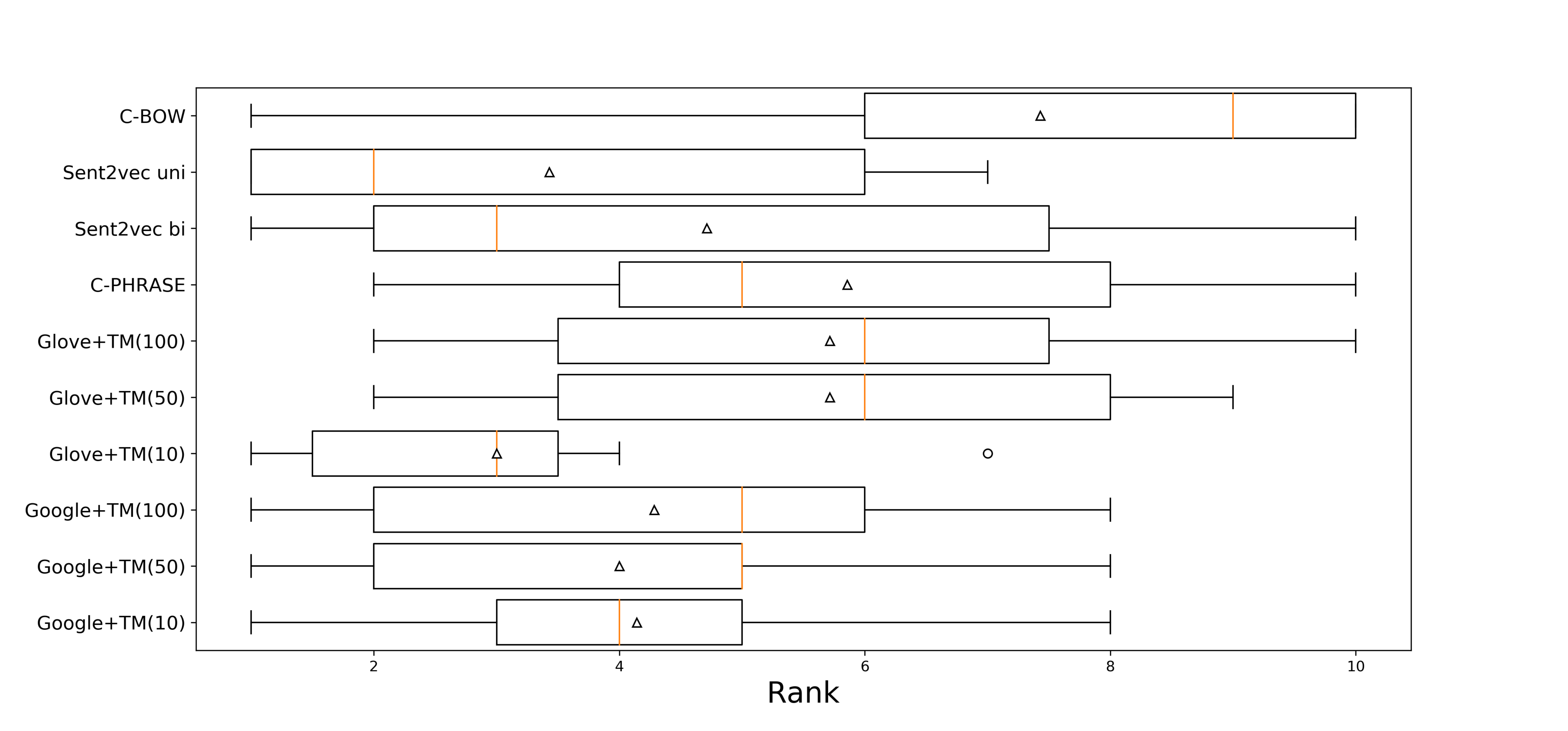} \\
			\footnotesize{(b) Spearman's $\rho$} \\			
	\end{tabular}} %
	\caption{Box plot of performance rank for each model}
	\label{fig:boxplot}%
\end{figure*}

The experimental results show that the proposed approach is ranked among the top three for most of the datasets despite its low computational complexity. We further compared the performance ranks of the proposed models with three selected benchmark models: Sent2vec, C-BOW, and C-PHRASE. The selection criteria are: (1) the model was tested for both STS 14 and SICK experiments, and (2) it performed the best for at least one dataset. For each dataset, we recorded the performance ranks of the models and then compared the distribution of the ranks and its average. Figure \ref{fig:boxplot} shows the box plots of the performance ranks for the selected ten models. The triangles denote the average performance ranks. The results show that the model which used Glove vectors and 10 \% of the total paraphrase sentences yielded the lowest average rank in both Pearson's $r$ and Spearman's $\rho$. In addition, in the case of Pearson's $r$, most of our proposed models resulted in lower average ranks than those of Sent2vec, which showed the best performance among the benchmark models.

\begin{table*}[t!]
	\begin{center}
		\caption{Detailed comparison with SIF} \label{table3.Compare_SIF}%
		\renewcommand{\arraystretch}{1.2}
		\centering{\setlength\tabcolsep{10pt}
			\begin{tabular}{*6c}
				\toprule
				{Model}  & {STS 12} & {STS 13} & {STS 14} & {STS 15} & {SICK} \\ \hline
				SIF (Glove + WR)& 56.2 & 56.6 & 68.5 & 71.7 & 72.2 \\
				Glove+TM(10) & 58.7 & \textbf{65.9} & 69.9 & 66.1 & \textbf{76.4} \\
				Goggle w2v+TM(50) & \textbf{59.2} & 65.2 & \textbf{69.8} & \textbf{72.4} & 73.9\\
				\bottomrule
		\end{tabular}}
	\end{center}
\end{table*}

\noindent \textbf{Comparison with SIF:} SIF has characteristics similar to our approach, i.e., it also computes a sentence representation vector from fixed pre-trained word vectors in a simple manner. In Table \ref{table2.Comparison}, we were not able to compare the performance with SIF in detail because \citet{SIF} did not specify the experimental result for the individual datasets of the STS 2014 task. However, because \citet{SIF} performed STS 2012--2015 tasks and the SICK semantic relatedness task, we compared the performance of our approach with SIF in more detail, as shown in Table \ref{table3.Compare_SIF}. We recorded the Pearson's $r$ of the best model for each pre-trained word vector. The experimental results reveal that our proposed approach showed a better performance than SIF for all the datasets. We also observed that the average performance improvement ratio was 6 \% and the performance was enhanced to a maximum of 16.5 \% compared to that of SIF.

\section{Conclusion}
Sentence embedding, which transforms an unstructured textual data into a structured vector form, is a fundamental and imperative method in the field of NLP. Generating sentence vectors that preserve the semantic meaning of a sentence is the key component to achieve an improved performance in various NLP tasks. Hence, various sentence embedding models have been proposed, which show an enhanced performance in various NLP tasks such as document classification, sentiment analysis, and semantic similarity task.

In this study, we first defined the property of semantic coherence that closely preserves the sentence semantics. Subsequently, we derived the objective function to train the transition matrix, which refines sentence representation vectors to satisfy the derived property. Finally, the transition matrix is trained by employing the MSCOCO caption dataset, which is widely used as a paraphrase dataset.

The proposed approach was evaluated through various semantic textual similarity tasks. Despite its low computational complexity, our approach showed significant improvements in various STS tasks. In addition, compared to the previously proposed benchmark models, our approach showed the best performance in many datasets and one of the best performances in almost every dataset.

Nonetheless, the proposed method has a limitation that paraphrase data is required to train the transition matrix. Although the experimental results show that the performance of a transition matrix is independent of the amount of training data, it can be a critical issue when analyzing the languages which are difficult to obtain paraphrase data. Therefore, similar to researches on unsupervised machine translation \cite{artetxe2017unsupervised,lample2017unsupervised,lample2018phrase} and unsupervised cross-lingual embedding \cite{wada2018unsupervised,xu2018unsupervised,conneau2017word}, both of which do not require labeled data, a research to generate sentence vectors preserving the latent meaning of sentences without paraphrase data should be developed.
	
\bibliographystyle{./sty/icml2017} 
\bibliography{References}

\begin{thebibliography}{43}
\providecommand{\natexlab}[1]{#1}
\providecommand{\url}[1]{\texttt{#1}}
\expandafter\ifx\csname urlstyle\endcsname\relax
  \providecommand{\doi}[1]{doi: #1}\else
  \providecommand{\doi}{doi: \begingroup \urlstyle{rm}\Url}\fi

\bibitem[Agirre et~al.(2012)Agirre, Diab, Cer, and Gonzalez-Agirre]{STS12}
Agirre, Eneko, Diab, Mona, Cer, Daniel, and Gonzalez-Agirre, Aitor.
\newblock Semeval-2012 task 6: A pilot on semantic textual similarity.
\newblock In \emph{Proceedings of the First Joint Conference on Lexical and
  Computational Semantics-Volume 1: Proceedings of the main conference and the
  shared task, and Volume 2: Proceedings of the Sixth International Workshop on
  Semantic Evaluation}, pp.\  385--393. Association for Computational
  Linguistics, 2012.

\bibitem[Agirre et~al.(2013)Agirre, Cer, Diab, Gonzalez-Agirre, and Guo]{STS13}
Agirre, Eneko, Cer, Daniel, Diab, Mona, Gonzalez-Agirre, Aitor, and Guo,
  Weiwei.
\newblock * sem 2013 shared task: Semantic textual similarity.
\newblock In \emph{Second Joint Conference on Lexical and Computational
  Semantics (* SEM), Volume 1: Proceedings of the Main Conference and the
  Shared Task: Semantic Textual Similarity}, volume~1, pp.\  32--43, 2013.

\bibitem[Agirre et~al.(2014)Agirre, Banea, Cardie, Cer, Diab, Gonzalez-Agirre,
  Guo, Mihalcea, Rigau, and Wiebe]{STS14}
Agirre, Eneko, Banea, Carmen, Cardie, Claire, Cer, Daniel, Diab, Mona,
  Gonzalez-Agirre, Aitor, Guo, Weiwei, Mihalcea, Rada, Rigau, German, and
  Wiebe, Janyce.
\newblock Semeval-2014 task 10: Multilingual semantic textual similarity.
\newblock In \emph{Proceedings of the 8th international workshop on semantic
  evaluation (SemEval 2014)}, pp.\  81--91, 2014.

\bibitem[Agirre et~al.(2015)Agirre, Banea, Cardie, Cer, Diab, Gonzalez-Agirre,
  Guo, Lopez-Gazpio, Maritxalar, Mihalcea, et~al.]{STS15}
Agirre, Eneko, Banea, Carmen, Cardie, Claire, Cer, Daniel, Diab, Mona,
  Gonzalez-Agirre, Aitor, Guo, Weiwei, Lopez-Gazpio, Inigo, Maritxalar, Montse,
  Mihalcea, Rada, et~al.
\newblock Semeval-2015 task 2: Semantic textual similarity, english, spanish
  and pilot on interpretability.
\newblock In \emph{Proceedings of the 9th international workshop on semantic
  evaluation (SemEval 2015)}, pp.\  252--263, 2015.

\bibitem[Agirre et~al.(2016)Agirre, Banea, Cer, Diab, Gonzalez-Agirre,
  Mihalcea, Rigau, and Wiebe]{STS2016}
Agirre, Eneko, Banea, Carmen, Cer, Daniel, Diab, Mona, Gonzalez-Agirre, Aitor,
  Mihalcea, Rada, Rigau, German, and Wiebe, Janyce.
\newblock Semeval-2016 task 1: Semantic textual similarity, monolingual and
  cross-lingual evaluation.
\newblock In \emph{Proceedings of the 10th International Workshop on Semantic
  Evaluation (SemEval-2016)}, pp.\  497--511, 2016.

\bibitem[Arora et~al.(2017)Arora, Liang, and Ma]{SIF}
Arora, Sanjeev, Liang, Yingyu, and Ma, Tengyu.
\newblock A simple but tough-to-beat baseline for sentence embeddings.
\newblock \emph{International conference on Learning Representations}, 2017.

\bibitem[Artetxe et~al.(2017)Artetxe, Labaka, Agirre, and
  Cho]{artetxe2017unsupervised}
Artetxe, Mikel, Labaka, Gorka, Agirre, Eneko, and Cho, Kyunghyun.
\newblock Unsupervised neural machine translation.
\newblock \emph{arXiv preprint arXiv:1710.11041}, 2017.

\bibitem[Bowman et~al.(2015)Bowman, Angeli, Potts, and Manning]{SNLI}
Bowman, Samuel~R, Angeli, Gabor, Potts, Christopher, and Manning,
  Christopher~D.
\newblock A large annotated corpus for learning natural language inference.
\newblock \emph{arXiv preprint arXiv:1508.05326}, 2015.

\bibitem[Conneau et~al.(2017{\natexlab{a}})Conneau, Kiela, Schwenk, Barrault,
  and Bordes]{InferSent}
Conneau, Alexis, Kiela, Douwe, Schwenk, Holger, Barrault, Loic, and Bordes,
  Antoine.
\newblock Supervised learning of universal sentence representations from
  natural language inference data.
\newblock \emph{arXiv preprint arXiv:1705.02364}, 2017{\natexlab{a}}.

\bibitem[Conneau et~al.(2017{\natexlab{b}})Conneau, Lample, Ranzato, Denoyer,
  and J{\'e}gou]{conneau2017word}
Conneau, Alexis, Lample, Guillaume, Ranzato, Marc'Aurelio, Denoyer, Ludovic,
  and J{\'e}gou, Herv{\'e}.
\newblock Word translation without parallel data.
\newblock \emph{arXiv preprint arXiv:1710.04087}, 2017{\natexlab{b}}.

\bibitem[Conneau et~al.(2017{\natexlab{c}})Conneau, Schwenk, Barrault, and
  Lecun]{conneau2017very}
Conneau, Alexis, Schwenk, Holger, Barrault, Lo{\"\i}c, and Lecun, Yann.
\newblock Very deep convolutional networks for text classification.
\newblock In \emph{Proceedings of the 15th Conference of the European Chapter
  of the Association for Computational Linguistics: Volume 1, Long Papers},
  volume~1, pp.\  1107--1116, 2017{\natexlab{c}}.

\bibitem[Deng et~al.(2009)Deng, Dong, Socher, Li, Li, and Fei-Fei]{Imagenet}
Deng, Jia, Dong, Wei, Socher, Richard, Li, Li-Jia, Li, Kai, and Fei-Fei, Li.
\newblock Imagenet: A large-scale hierarchical image database.
\newblock In \emph{Computer Vision and Pattern Recognition, 2009. CVPR 2009.
  IEEE Conference on}, pp.\  248--255. IEEE, 2009.

\bibitem[Glorot \& Bengio(2010)Glorot and Bengio]{Xavier}
Glorot, Xavier and Bengio, Yoshua.
\newblock Understanding the difficulty of training deep feedforward neural
  networks.
\newblock In \emph{Proceedings of the Thirteenth International Conference on
  Artificial Intelligence and Statistics}, pp.\  249--256, 2010.

\bibitem[Gupta et~al.(2017)Gupta, Agarwal, Singh, and Rai]{gupta2017deep}
Gupta, Ankush, Agarwal, Arvind, Singh, Prawaan, and Rai, Piyush.
\newblock A deep generative framework for paraphrase generation.
\newblock \emph{arXiv preprint arXiv:1709.05074}, 2017.

\bibitem[Hill et~al.(2016{\natexlab{a}})Hill, Cho, and Korhonen]{FastSent}
Hill, Felix, Cho, Kyunghyun, and Korhonen, Anna.
\newblock Learning distributed representations of sentences from unlabelled
  data.
\newblock \emph{arXiv preprint arXiv:1602.03483}, 2016{\natexlab{a}}.

\bibitem[Hill et~al.(2016{\natexlab{b}})Hill, Cho, and
  Korhonen]{hill2016learning}
Hill, Felix, Cho, Kyunghyun, and Korhonen, Anna.
\newblock Learning distributed representations of sentences from unlabelled
  data.
\newblock \emph{arXiv preprint arXiv:1602.03483}, 2016{\natexlab{b}}.

\bibitem[Hochreiter \& Schmidhuber(1997)Hochreiter and Schmidhuber]{LSTM}
Hochreiter, Sepp and Schmidhuber, J{\"u}rgen.
\newblock Long short-term memory.
\newblock \emph{Neural computation}, 9\penalty0 (8):\penalty0 1735--1780, 1997.

\bibitem[Hu et~al.(2014)Hu, Lu, Li, and Chen]{hu2014convolutional}
Hu, Baotian, Lu, Zhengdong, Li, Hang, and Chen, Qingcai.
\newblock Convolutional neural network architectures for matching natural
  language sentences.
\newblock In \emph{Advances in neural information processing systems}, pp.\
  2042--2050, 2014.

\bibitem[Jang \& Kang(2018)Jang and Kang]{jang2018paraphrase}
Jang, Myeongjun and Kang, Pilsung.
\newblock Paraphrase thought: Sentence embedding module imitating human
  language recognition.
\newblock \emph{arXiv preprint arXiv:1808.05505}, 2018.

\bibitem[Kenter et~al.(2016)Kenter, Borisov, and de~Rijke]{kenter2016siamese}
Kenter, Tom, Borisov, Alexey, and de~Rijke, Maarten.
\newblock Siamese cbow: Optimizing word embeddings for sentence
  representations.
\newblock \emph{arXiv preprint arXiv:1606.04640}, 2016.

\bibitem[Kim(2014)]{kim2014convolutional}
Kim, Yoon.
\newblock Convolutional neural networks for sentence classification.
\newblock \emph{arXiv preprint arXiv:1408.5882}, 2014.

\bibitem[Kiros et~al.(2015)Kiros, Zhu, Salakhutdinov, Zemel, Urtasun, Torralba,
  and Fidler]{SkipThought}
Kiros, Ryan, Zhu, Yukun, Salakhutdinov, Ruslan~R, Zemel, Richard, Urtasun,
  Raquel, Torralba, Antonio, and Fidler, Sanja.
\newblock Skip-thought vectors.
\newblock In \emph{Advances in neural information processing systems}, pp.\
  3294--3302, 2015.

\bibitem[Koch(2015)]{koch2015siamese}
Koch, Gregory.
\newblock Siamese neural networks for one-shot image recognition.
\newblock In \emph{ICML Deep Learning workshop}, volume~2, 2015.

\bibitem[Kruszewski et~al.(2015)Kruszewski, Lazaridou, Baroni,
  et~al.]{kruszewski2015jointly}
Kruszewski, Germ{\'a}n, Lazaridou, Angeliki, Baroni, Marco, et~al.
\newblock Jointly optimizing word representations for lexical and sentential
  tasks with the c-phrase model.
\newblock In \emph{Proceedings of the 53rd Annual Meeting of the Association
  for Computational Linguistics and the 7th International Joint Conference on
  Natural Language Processing (Volume 1: Long Papers)}, volume~1, pp.\
  971--981, 2015.

\bibitem[Lample et~al.(2017)Lample, Denoyer, and
  Ranzato]{lample2017unsupervised}
Lample, Guillaume, Denoyer, Ludovic, and Ranzato, Marc'Aurelio.
\newblock Unsupervised machine translation using monolingual corpora only.
\newblock \emph{arXiv preprint arXiv:1711.00043}, 2017.

\bibitem[Lample et~al.(2018)Lample, Ott, Conneau, Denoyer, and
  Ranzato]{lample2018phrase}
Lample, Guillaume, Ott, Myle, Conneau, Alexis, Denoyer, Ludovic, and Ranzato,
  Marc'Aurelio.
\newblock Phrase-based \& neural unsupervised machine translation.
\newblock \emph{arXiv preprint arXiv:1804.07755}, 2018.

\bibitem[Le \& Mikolov(2014)Le and Mikolov]{Doc2vec}
Le, Quoc and Mikolov, Tomas.
\newblock Distributed representations of sentences and documents.
\newblock In \emph{International Conference on Machine Learning}, pp.\
  1188--1196, 2014.

\bibitem[Lin et~al.(2014)Lin, Maire, Belongie, Hays, Perona, Ramanan,
  Doll{\'a}r, and Zitnick]{MSCOCO}
Lin, Tsung-Yi, Maire, Michael, Belongie, Serge, Hays, James, Perona, Pietro,
  Ramanan, Deva, Doll{\'a}r, Piotr, and Zitnick, C~Lawrence.
\newblock Microsoft coco: Common objects in context.
\newblock In \emph{European conference on computer vision}, pp.\  740--755.
  Springer, 2014.

\bibitem[Marelli et~al.(2014)Marelli, Bentivogli, Baroni, Bernardi, Menini, and
  Zamparelli]{marelli2014semeval}
Marelli, Marco, Bentivogli, Luisa, Baroni, Marco, Bernardi, Raffaella, Menini,
  Stefano, and Zamparelli, Roberto.
\newblock Semeval-2014 task 1: Evaluation of compositional distributional
  semantic models on full sentences through semantic relatedness and textual
  entailment.
\newblock In \emph{Proceedings of the 8th international workshop on semantic
  evaluation (SemEval 2014)}, pp.\  1--8, 2014.

\bibitem[Mikolov et~al.(2013{\natexlab{a}})Mikolov, Le, and
  Sutskever]{mikolov2013exploiting}
Mikolov, Tomas, Le, Quoc~V, and Sutskever, Ilya.
\newblock Exploiting similarities among languages for machine translation.
\newblock \emph{arXiv preprint arXiv:1309.4168}, 2013{\natexlab{a}}.

\bibitem[Mikolov et~al.(2013{\natexlab{b}})Mikolov, Sutskever, Chen, Corrado,
  and Dean]{Word2vec}
Mikolov, Tomas, Sutskever, Ilya, Chen, Kai, Corrado, Greg~S, and Dean, Jeff.
\newblock Distributed representations of words and phrases and their
  compositionality.
\newblock In \emph{Advances in neural information processing systems}, pp.\
  3111--3119, 2013{\natexlab{b}}.

\bibitem[Pagliardini et~al.(2017)Pagliardini, Gupta, and Jaggi]{Sent2vec}
Pagliardini, Matteo, Gupta, Prakhar, and Jaggi, Martin.
\newblock Unsupervised learning of sentence embeddings using compositional
  n-gram features.
\newblock \emph{arXiv preprint arXiv:1703.02507}, 2017.

\bibitem[Pearson(1895)]{pearson1895note}
Pearson, Karl.
\newblock Note on regression and inheritance in the case of two parents.
\newblock \emph{Proceedings of the Royal Society of London}, 58:\penalty0
  240--242, 1895.

\bibitem[Pennington et~al.(2014)Pennington, Socher, and Manning]{Glove}
Pennington, Jeffrey, Socher, Richard, and Manning, Christopher.
\newblock Glove: Global vectors for word representation.
\newblock In \emph{Proceedings of the 2014 conference on empirical methods in
  natural language processing (EMNLP)}, pp.\  1532--1543, 2014.

\bibitem[Prakash et~al.(2016)Prakash, Hasan, Lee, Datla, Qadir, Liu, and
  Farri]{prakash2016neural}
Prakash, Aaditya, Hasan, Sadid~A, Lee, Kathy, Datla, Vivek, Qadir, Ashequl,
  Liu, Joey, and Farri, Oladimeji.
\newblock Neural paraphrase generation with stacked residual lstm networks.
\newblock \emph{arXiv preprint arXiv:1610.03098}, 2016.

\bibitem[Smith et~al.(2017)Smith, Turban, Hamblin, and
  Hammerla]{smith2017offline}
Smith, Samuel~L, Turban, David~HP, Hamblin, Steven, and Hammerla, Nils~Y.
\newblock Offline bilingual word vectors, orthogonal transformations and the
  inverted softmax.
\newblock \emph{arXiv preprint arXiv:1702.03859}, 2017.

\bibitem[Spearman(1904)]{spearman1904proof}
Spearman, Charles.
\newblock The proof and measurement of association between two things.
\newblock \emph{The American journal of psychology}, 15\penalty0 (1):\penalty0
  72--101, 1904.

\bibitem[Sutskever et~al.(2014)Sutskever, Vinyals, and
  Le]{sutskever2014sequence}
Sutskever, Ilya, Vinyals, Oriol, and Le, Quoc~V.
\newblock Sequence to sequence learning with neural networks.
\newblock In \emph{Advances in neural information processing systems}, pp.\
  3104--3112, 2014.

\bibitem[Wada \& Iwata(2018)Wada and Iwata]{wada2018unsupervised}
Wada, Takashi and Iwata, Tomoharu.
\newblock Unsupervised cross-lingual word embedding by multilingual neural
  language models.
\newblock \emph{arXiv preprint arXiv:1809.02306}, 2018.

\bibitem[Wan et~al.(2016)Wan, Lan, Guo, Xu, Pang, and Cheng]{wan2016deep}
Wan, Shengxian, Lan, Yanyan, Guo, Jiafeng, Xu, Jun, Pang, Liang, and Cheng,
  Xueqi.
\newblock A deep architecture for semantic matching with multiple positional
  sentence representations.
\newblock In \emph{AAAI}, volume~16, pp.\  2835--2841, 2016.

\bibitem[Wu et~al.(2016)Wu, Schuster, Chen, Le, Norouzi, Macherey, Krikun, Cao,
  Gao, Macherey, et~al.]{wu2016google}
Wu, Yonghui, Schuster, Mike, Chen, Zhifeng, Le, Quoc~V, Norouzi, Mohammad,
  Macherey, Wolfgang, Krikun, Maxim, Cao, Yuan, Gao, Qin, Macherey, Klaus,
  et~al.
\newblock Google's neural machine translation system: Bridging the gap between
  human and machine translation.
\newblock \emph{arXiv preprint arXiv:1609.08144}, 2016.

\bibitem[Xu et~al.(2018)Xu, Yang, Otani, and Wu]{xu2018unsupervised}
Xu, Ruochen, Yang, Yiming, Otani, Naoki, and Wu, Yuexin.
\newblock Unsupervised cross-lingual transfer of word embedding spaces.
\newblock \emph{arXiv preprint arXiv:1809.03633}, 2018.

\bibitem[Zhu et~al.(2015)Zhu, Kiros, Zemel, Salakhutdinov, Urtasun, Torralba,
  and Fidler]{Bookcorpus}
Zhu, Yukun, Kiros, Ryan, Zemel, Rich, Salakhutdinov, Ruslan, Urtasun, Raquel,
  Torralba, Antonio, and Fidler, Sanja.
\newblock Aligning books and movies: Towards story-like visual explanations by
  watching movies and reading books.
\newblock In \emph{Proceedings of the IEEE international conference on computer
  vision}, pp.\  19--27, 2015.

\end{thebibliography}
\end{CJK}
\end{document}